# Leveraging LLMs for Title and Abstract Screening for Systematic Review: A Cost-Effective Dynamic Few-Shot Learning Approach

Yun-Chung Liu[1,2†], Rui Yang[3†], Jonathan Chong Kai Liew[4], Ziran Yin[1], Henry Foote, MD[5], Christopher J. Lindsell, PhD[1,6], Chuan Hong, PhD[1,6*]

[1] *Department of Biostatistics and Bioinformatics, Duke University, Durham, NC, USA*

[2] *Interdisciplinary Data Science, Duke University, Durham, NC, USA*

[3] *Department of Biomedical Informatics, National University of Singapore, Singapore, Singapore*

[4] *School of Chemistry, Nanyang Technological University, Singapore, Singapore*

[5] *Department of Pediatrics, Duke University School of Medicine, Durham, NC, USA*

[6] *Duke Clinical Research Institute, Duke University School of Medicine, Durham, NC, USA*

[†]Yun-Chung Liu and Rui Yang are equally contributed

[*]Correspondence: Chuan Hong

Email: chuan.hong@duke.edu

## Abstract

Systematic reviews are a key component of evidence-based medicine, playing a critical role in synthesizing existing research evidence and guiding clinical decisions. However, with the rapid growth of research publications, conducting systematic reviews has become increasingly burdensome, with title and abstract screening being one of the most time-consuming and resource-intensive steps. To mitigate this issue, we designed a two-stage dynamic few-shot learning (DFSL) approach aimed at improving the efficiency and performance of large language models (LLMs) in the title and abstract screening task. Specifically, this approach first uses a low-cost LLM for initial screening, then re-evaluates low-confidence instances using a high-performance LLM, thereby enhancing screening performance while controlling computational costs. We evaluated this approach across 10 systematic reviews, and the results demonstrate its strong generalizability and cost-effectiveness, with potential to reduce manual screening burden and accelerate the systematic review process in practical applications.

## 1. Introduction

Systematic reviews are typically recognized as a key component of evidence-based medicine, playing a critical role in synthesizing existing research evidence and guiding clinical decisions [1,2]. However, with the rapid increase in research publications [3], conducting systematic reviews has become time-consuming [4]. A typical systematic review involves several key steps: identifying research objectives, defining inclusion/exclusion criteria, searching databases, screening titles and abstracts, reviewing full texts, extracting data, and synthesizing results [5]. Among these, the title and abstract screening stage alone can demand substantial resources to review thousands of publications [6].

To reduce the manual screening burden in systematic reviews, researchers have explored different automated screening methods. Early studies combined clinical term frequency distribution with word embeddings to identify studies aligned with research objectives [7]. Subsequent studies directly leveraged word embedding techniques to convert abstracts into embedding vectors and classify relevant studies [8]. Another study employed sentence transformers for training, performing contrastive learning by pairing included and excluded samples, enabling the model to learn the similarity features between them, thereby assisting in screening [9].

Recently, large language models (LLMs) have demonstrated exceptional capabilities in text understanding and generation [10,11] and have been explored for application to the title and abstract screening task in systematic reviews. Existing studies have explored zero-shot settings by designing prompts to guide LLMs in identifying relevant studies directly [12,13]. Additionally, chain-of-thought techniques have been introduced to enhance LLMs' reasoning capabilities in the screening task by prompting them to perform step-by-step analysis [14]. Others have adopted multi-agent frameworks to simulate real decision-making processes, where agents discuss inclusion and exclusion criteria, and reach consensus [15]. As LLMs continue to advance, their potential in assisting systematic review screening grows, promising to reduce researchers' workload and improve efficiency [2].

However, despite their demonstrated potential, LLM-driven screening approaches may exhibit inconsistent performance when applied to diverse medical topics with varying data distributions [13]. Additionally, the computational cost associated with deploying high-performance LLMs can be substantial, making it impractical for large-scale systematic reviews without cost-effective strategies. This highlights the need for a generalizable and economically feasible approach that can effectively balance screening performance and resource allocation.

To address the dual challenge of performance and cost in large-scale title and abstract screening across diverse medical topics, we developed a dynamic few-shot learning (DFSL) approach designed to efficiently integrate into the systematic review process. The DFSL approach adopts a two-stage structure: the first stage performs initial screening using a low-cost LLM guided by dynamically constructed few-shot prompts and provides confidence scores; the second stage re-evaluates low-confidence instances using a high-performance LLM. This approach achieves an effective balance between screening performance and computational resources by combining adaptive prompting with the confidence-driven triage mechanism. We evaluated this approach on ten systematic review datasets covering different stages including diagnostic, intervention, and prognostic studies, encompassing nearly 10,000 studies [16].

## 2. Methods

In the Methods section, we first describe the standard process of systematic reviews. Subsequently, we provide a detailed description of the data curation process of this work, benchmarking of different prompting strategies, and the design of the DFSL approach, which includes a cost-effective two-stage screening strategy that achieves dual improvements in both performance and efficiency through the adoption of dynamic few-shot prompting and the confidence-driven LLM selection mechanism.

## 2.1 Standard Systematic Review Workflow

Systematic reviews follow a structured, multi-step process to ensure comprehensive and unbiased evidence synthesis [5]. Key steps typically include: (1) formulating a research objective; (2) defining inclusion and exclusion criteria; (3) conducting systematic searches across databases; (4) screening titles and abstracts for relevance; (5) reviewing full texts; (6) extracting data; and (7) synthesizing results through qualitative or quantitative analysis. Among these, title and abstract screening is particularly time-consuming and resource-intensive, as it involves manually screening thousands of studies to determine eligibility based on predefined criteria. This step serves as a critical bottleneck and is the primary focus of the DFSL approach developed in this study.

## 2.2 Data Curation

We utilized the CLEF 2019 dataset, which includes 31 systematic reviews covering diagnosis, intervention, and prognosis, each providing PMIDs of all studies retrieved [14]. To evaluate our framework, we selected a subset of 10 systematic reviews (5 diagnostic reviews, 4 intervention reviews, and 1 prognostic review) [15-24]. The selection criteria were systematic reviews with more than 100 studies retrieved during the search stage and a reasonable inclusion ratio. This criterion is crucial as a small sample size or low inclusion ratio could result in unstable evaluation outcomes with limited statistical significance. Using the provided PMIDs, we retrieved the corresponding titles and abstracts for each study through the PubMed API, excluding records with missing content and subsequently performing deduplication to construct the final evaluation dataset. The inclusion and exclusion criteria were directly obtained from the original systematic reviews.

## 2.3 Benchmarking Prompting Strategies

To ensure fair comparison, all methods (zero-shot, chain-of-thought, few-shot, and DFSL) used the same model with identical settings. The only difference across methods lies in the prompt construction strategy itself.

#### *2.3.1 Zero-Shot Learning*

Zero-shot learning refers to an LLM's ability to produce outputs without any training instances [10, 11]. In the systematic review screening task, we directly input the title, abstract, and inclusion/exclusion criteria into the LLM, instructing it to output either "include" or "exclude" without providing any instances with labels.

#### *2.3.2 Chain-of-Thought*

Chain-of-thought is a strategy that guides LLMs through a step-by-step reasoning process [27]. Unlike directly generating predictions, this approach requires the LLM to produce a sequence of logical steps before arriving at the final decision. In the context of abstract screening, we extended the zero-shot setting by adding instructions such as "think step by step", encouraging the LLM to reason through the decision-making process.

#### *2.3.3 Few-Shot Learning*

Few-shot learning guides LLMs to learn specific tasks by providing a small number of labeled examples [28]. In the title and abstract screening task, we randomly provided several known instances of titles and abstracts that should be included or excluded, serving as references for the LLM's screening process.

#### *2.3.4 DFSL: A Cost-Effective Dynamic Few-Shot Learning Screening Approach*

Inspired by prior work on dynamic instance selection [29], we propose the DFSL approach. This approach first clustered studies and selected representative inclusion and exclusion instances from each cluster to construct an instance pool. Subsequently, for each study to be classified, the most similar instances were dynamically selected from the instance pool based on its cluster assignment, thereby constructing dynamic few-shot prompts. We systematically applied this approach to the title and abstract screening task in systematic reviews. Figure 1 presents an overview of the entire pipeline.

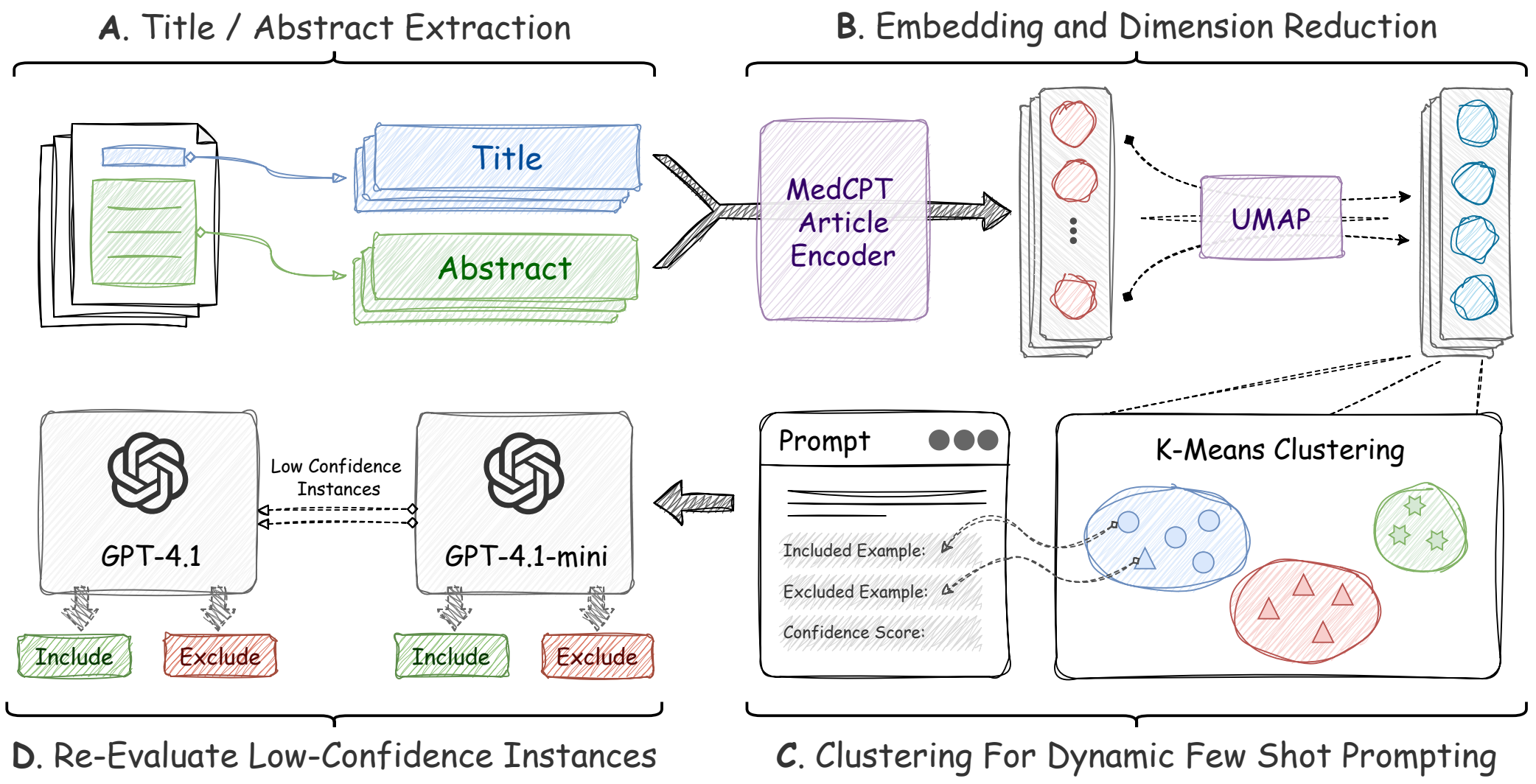

**Figure 1 Overview of the entire title and abstract screening pipeline.** The pipeline includes: (A) extracting titles and abstracts from systematic reviews; (B) generating text embeddings and performing dimensionality reduction; (C) conducting clustering, dynamically constructing few-shot prompts, and performing initial screening with confidence scores; (D) using a high-performance LLM to re-evaluate low-confidence instances (identified based on confidence scores below the threshold of 0.9 in the initial screening stage, where confidence scores range from 0 to 1).

Specifically, we utilized "MedCPT Article Encoder" [30] to generate text embeddings for all studies, followed by dimensionality reduction to 2 dimensions through the UMAP [31] algorithm. Subsequently, we applied the K-means algorithm [32] for clustering and, according to the data characteristics, selected one inclusion instance and two exclusion instances from each cluster to construct dynamic few-shot prompts. This approach not only addresses the randomness problem in traditional few-shot learning instance selection but also optimizes screening performance across diverse medical topics. Additionally, the DFSL approach supports confidence evaluation by requiring the LLM to generate confidence scores ranging from 0 to 1 for each prediction, allowing further optimization of low-confidence instances. The evaluation prompt for the DFSL approach is provided in Supplementary Information A.

Building on this, to enhance screening efficiency while reducing computational costs, we implemented a cost-effective two-stage screening strategy. In the first stage, GPT-4.1-mini was employed for the initial screening of all titles and abstracts, with input costs of $0.40/1M tokens and output costs of $1.60/1M tokens, thereby generating preliminary screening results at a lower cost. Simultaneously, the model was required to provide a confidence score ranging from 0 to 1 for each prediction, reflecting the degree of certainty in its current judgment.

Based on the confidence scores generated, we set 0.9 as the threshold for identifying low-confidence instances, which were then further processed in the second stage using the higher-performance GPT-4.1, with input costs of $2.00/1M tokens and output costs of $8.00/1M tokens. The selection of this threshold comprehensively considers the trade-off between confidence score distribution and computational costs: when the confidence score threshold is set too low, only a very small number of instances would be sent to the second stage, making it difficult to fully leverage the role of the high-performance LLM in correcting potential misclassifications; conversely, when the threshold is set too high, the proportion of instances requiring re-evaluation significantly increases, resulting in higher computational costs and weakening the cost advantage of the two-stage strategy. After considering the re-evaluation ratio, cost overhead, and overall performance under different thresholds, we found that 0.9 provides a relatively stable and reasonable trade-off point between screening effectiveness and resource consumption. We provided sensitivity analysis of different confidence score thresholds in Supplementary Information B.

By assigning the majority of screening instances to a low-cost LLM and selectively invoking a high-performance LLM only for low-confidence, potentially uncertain instances, this approach can effectively balance screening performance and cost-efficiency.

## 3. Results

### 3.1 Data characteristics

Table 1 shows the characteristics of the ten included systematic reviews from CLEF 19, published between 2014 and 2021, covering diagnostic, intervention, and prognostic stages. According to the original study, the total number of titles and abstracts generated from the search step was 10,320. After data curation, the actual number of titles and abstracts included was 9,515, of which 1,052 were identified as ones should be included in their corresponding systematic reviews during the title and abstract screening stage. The range of included abstracts was from 119 to 3,344 and the average inclusion rate was 11.1%. Different numbers of clusters were assigned to the systematic reviews based on the number of titles and abstracts in each systematic review. In the DFSL approach, one positive instance and two negative instances were used for each cluster.

**Table 1 Number of titles, abstracts, and included studies across ten systematic reviews.**

| | Stage | Retrieved Records | Curated Records | Clusters |
|---|---|---|---|---|
| *CD012233* | Diagnosis | 472 (43) | 429 (38) | 5 |
| *CD012768* | | 131 (45) | 119 (43) | 3 |
| *CD011977* | | 1182 (297) | 1117 (280) | 8 |
| *CD012069* | | 251 (42) | 247 (41) | 4 |
| *CD012551* | | 316 (47) | 307 (46) | 4 |
| *CD004414* | Intervention | 195 (49) | 192 (49) | 3 |
| *CD012661* | | 3479 (320) | 2897 (282) | 10 |
| *CD011431* | | 591 (68) | 546 (65) | 5 |
| *CD011420* | | 336 (16) | 303 (16) | 4 |
| *CD010772* | Prognosis | 3367 (192) | 3343 (192) | 10 |

### 3.2 Performance of the DFSL approach

As shown in Figure 2, our approach demonstrated strong overall performance in the title and abstract screening task, achieving significant improvements across all ten systematic reviews, with an average F1 score of 0.552, clearly outperforming all baseline methods—zero-shot learning (0.458), chain-of-thought (0.458), and few-shot learning (0.508). The advantage of DFSL lies primarily in its dynamic instance selection strategy. Unlike traditional methods that rely on randomly or statically selected instances, DFSL dynamically selects the most relevant positive and negative instances for each study to be classified, and this targeted prompting helps guide the LLM to make more accurate predictions. Additionally, we provide more evaluation metrics in Supplementary Information C, including precision, recall, and accuracy.

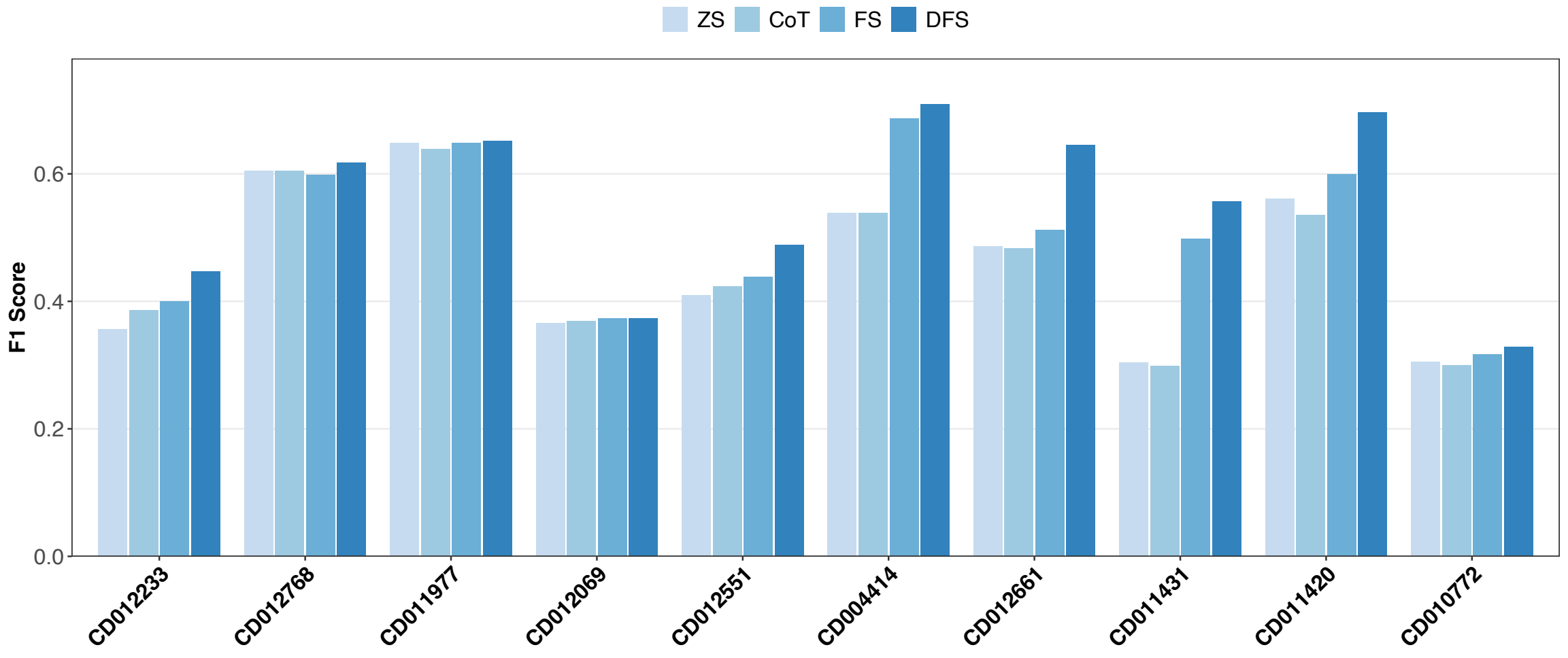

**Figure 2 F1 score performance comparison of different prompting strategies (Zero-Shot (ZS), Chain-of-Thought (CoT), Few-Shot (FS), Dynamic Few-Shot (DFS)) across ten systematic reviews.** DFS achieved higher F1 scores in the title and abstract screening task across all ten systematic reviews compared to Zero-Shot, Chain-of-Thought, and Few-Shot methods.

### 3.3 Performance comparison with open-weight LLMs

In our DFSL approach, we adopted the proprietary LLM as the default configuration for the initial screening stage. To provide a more comprehensive comparative analysis, we further evaluated several representative open-weight general and medical-specific LLMs under the zero-shot setting, including Phi-3.5-mini [33], MediPhi [34], Gemma3-4B [35], Med-Gemma-4B [36], and Med-Gemma-27B [36]. As shown in Table 2, the evaluation results showed that Gemma3-4B and GPT-4.1-mini demonstrated superior performance on certain systematic reviews, outperforming other LLMs. Notably, medical-specific LLMs did not exhibit any advantages over their corresponding general LLMs.

**Table 2 F1 score comparison with open-weight LLMs under the zero-shot setting.**

| | GPT-4.1-mini | Phi-3.5-mini | MediPhi | Gemma3-4b | MedGemma-4b | MedGemma-27b |
|---|---|---|---|---|---|---|
| *CD012233* | **0.3571** | 0.2396 | 0.2444 | 0.2975 | 0.2376 | 0.3529 |
| *CD012768* | 0.6055 | 0.5455 | 0.3492 | **0.6387** | 0.5825 | 0.6218 |
| *CD011977* | 0.6486 | 0.5745 | 0.2222 | **0.6972** | 0.6596 | 0.6076 |
| *CD012069* | **0.3662** | 0.2375 | 0.1579 | 0.2635 | 0.2670 | 0.3152 |
| *CD012551* | 0.4094 | 0.2907 | 0.4021 | 0.3152 | 0.3969 | **0.4267** |
| *CD004414* | 0.5385 | 0.3860 | 0.2632 | 0.4062 | 0.5161 | **0.6486** |
| *CD012661* | **0.4871** | 0.3863 | 0.1273 | 0.3546 | 0.3667 | 0.4403 |
| *CD011431* | 0.3045 | 0.5642 | 0.4140 | **0.6278** | 0.5520 | 0.1742 |
| *CD011420* | 0.5614 | 0.5581 | 0.4000 | **0.7105** | 0.6667 | 0.6462 |
| *CD010772* | 0.3051 | 0.5225 | 0.2182 | **0.5962** | 0.5275 | 0.2667 |

Different LLMs exhibit significant disparities in deployment methods, hardware requirements, and operational efficiency, factors that hold important practical implications in the actual large-scale literature screening scenario. It should be emphasized that the DFSL approach itself does not depend on any specific LLM; its framework design allows users to flexibly select and integrate different proprietary or open-weight LLMs into the screening stage based on actual resource conditions and application requirements.

### 3.3 Effectiveness of re-evaluating low-confidence instances

In the DFSL approach, all titles and abstracts were first screened initially by the low-cost GPT-4.1-mini; for studies with confidence scores below 0.9, they proceeded to the second stage for re-evaluation by the higher-performance GPT-4.1. This two-stage strategy achieved stable performance improvements while controlling costs. As shown in Figure 3, the re-evaluation stage resulted in improvements in 9 out of 10 systematic reviews, with the overall F1 score increasing from 0.552 in the initial screening to 0.563. To verify the statistical significance of this improvement, we further conducted paired statistical tests: the paired t-test yielded $p = 0.008$, indicating that the improvements brought by the two-stage strategy were statistically significant. Notably, only 12.96% of the studies were identified as low-confidence and proceeded to the GPT-4.1 re-evaluation stage. In the re-evaluation of ten systematic reviews comprising over 9,500 records, the cost was only approximately $4, while the majority of the remaining screening work was completed by low-cost LLMs. This

design significantly reduced reliance on high-cost LLMs while maintaining high screening performance, thereby achieving a better cost-performance balance.

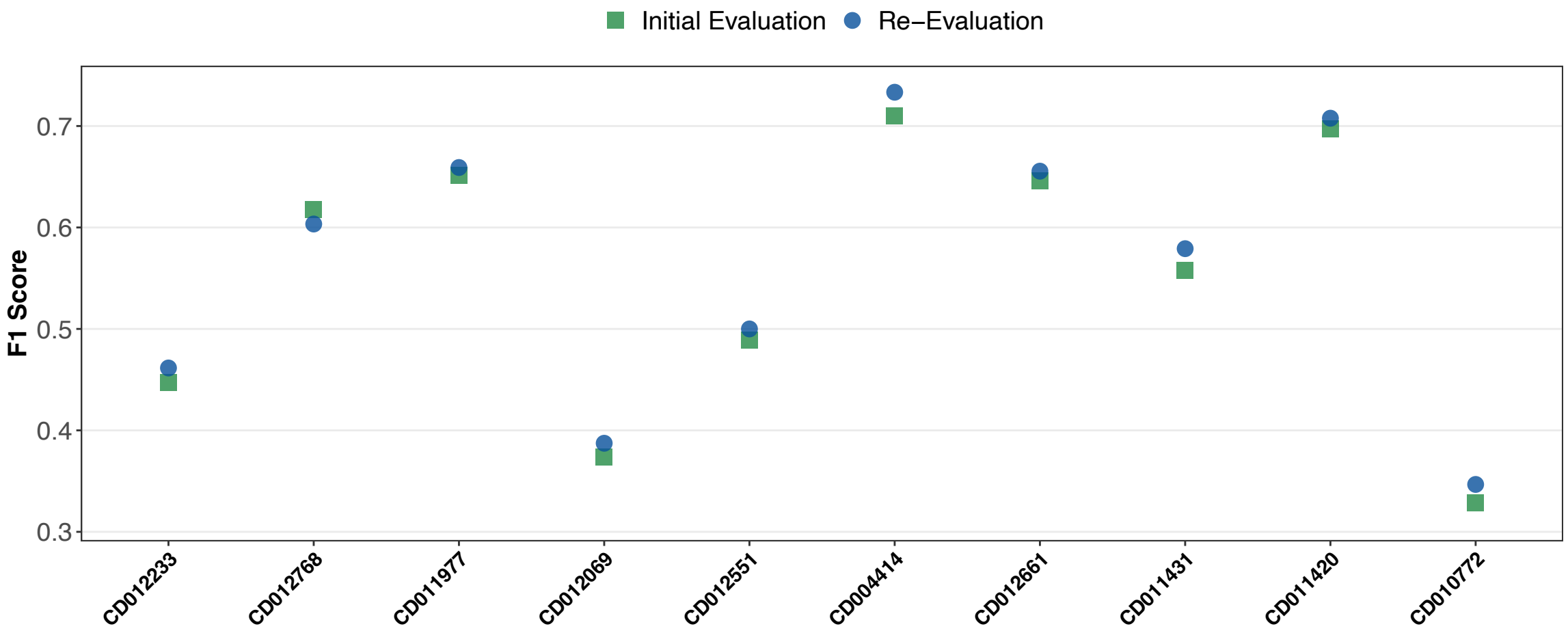

**Figure 3 F1 score performance before and after re-evaluation of low-confidence instances across ten systematic reviews.** The DFSL approach initially screened all studies using GPT-4.1-mini, and instances with confidence scores below 0.9 were re-evaluated using GPT-4.1.

## 4. Discussion

This study proposes the DFSL approach, aimed at improving LLM performance in the title and abstract screening task for systematic reviews. Through dynamic prompt construction, the DFSL approach demonstrated strong performance. Across ten systematic reviews spanning diverse stages, DFSL achieved an average F1 score of 0.552, substantially outperforming baseline prompting methods such as zero-shot, chain-of-thought, and few-shot.

DFSL offers advantages not only in performance but also in cost-effectiveness. By introducing the confidence score threshold, only predictions with the confidence score below 0.9 were re-evaluated using a higher-performance LLM. In our experiments, only 12.96% of instances required second-stage evaluation by GPT-4.1, yet the overall F1 score increased to 0.563 and passed statistical testing. This two-stage approach effectively balanced performance with resource usage, demonstrating scalability for large-scale literature screening applications. It is worth noting that, although this study employed GPT-4.1 in the second stage, the approach itself supports the flexible selection of more advanced models as needed, such as substantially more expensive reasoning-enhanced LLMs. In such scenarios, compared with uniformly applying high-performance LLMs to all studies, this approach is expected to provide significant cost savings.

In terms of applicability, DFSL showed consistent performance across various stages, including diagnosis, intervention, and prognosis, highlighting its generalizability. Since the approach does not require task-specific fine-tuning, it exhibits strong transferability and can be seamlessly integrated into existing systematic review workflows to assist in initial screening and improve efficiency. Additionally, the DFSL approach holds practical significance for real-world evidence synthesis. In the context of rapidly growing literature and high manual screening costs, it serves as a valuable solution to help researchers prioritize highly relevant studies, reduce manual workload, and accelerate the systematic review process.

Despite these encouraging results, several limitations remain. First, due to data access constraints and computational costs, the scale of this study was relatively limited; evaluating the approach on larger-scale datasets would help further validate its robustness. Second, although the approach improved performance, false negatives still occurred. Therefore, human review remains necessary in high-stakes settings, and the approach should be seen as an assistive tool rather than a complete replacement for expert judgment. Additionally, this study has not yet conducted human-in-the-loop evaluations, nor has it been integrated with real-world systematic review platforms such as Rayyan and ASReview or undergone user-level workflow validation. Future user studies conducted in practical usage scenarios would better demonstrate the potential value and feasibility of this approach in real-world practice.

## Data And Code Availability

All data used in this study were obtained from publicly available datasets. The code used in this study is available upon request.

## Author Contributions

Conceptualization, YCL, RY, and CH; Methodology, YCL, RY, and CH; Investigation, YCL, RY, JCKL, ZY, HF, and CH; Visualization, YCL, RY, ZY; Project administration, CH; Supervision, CH; Writing—Original Draft, YCL, RY, ZY; Writing—Review & Editing, RY, NL, CJL, and CH.

## Completing Interests

The authors do not have conflicts of interest related to this study. This research received no specific grant from any funding agency in the public, commercial or not-for-profit sectors.

## Ethical Approval

Not applicable.

## Clinical Trial Number

Not applicable.

## Consent to Participate

Not applicable.

## Funding Declaration

Not applicable.

# Supplementary Information

## Supplementary Information A

## Evaluation Prompt

> You are a reviewer conducting abstract screening for a systematic review. Your task is to determine whether the given title and abstract should be included or excluded based on the provided criteria.
>
> Only return the result: include or exclude.
>
> Criteria:
> {criteria}
>
> Title:
> {title}
>
> Abstract:
> {abstract}
>
> Output:

**Table 1 Zero-Shot Evaluation Prompt.**

> You are a reviewer conducting abstract screening for a systematic review. Your task is to determine whether the given title and abstract should be included or excluded based on the provided criteria. Think step by step.
>
> Only return the result: include or exclude.
>
> Criteria:
> {criteria}
>
> Title:
> {title}
>
> Abstract:
> {abstract}
>
> Output:

**Table 2 Chain of Thought Evaluation Prompt.**

You are a reviewer conducting abstract screening for a systematic review. Your task is to determine whether the given title and abstract should be included or excluded based on the provided criteria and examples.

Only return the result: include or exclude.

Criteria:
{criteria}

Instances:
{instances}

Title:
{title}

Abstract:
{abstract}

Output:

**Table 3 Few-Shot Evaluation Prompt.**

You are a reviewer conducting abstract screening for a systematic review. Your task is to determine whether the given title and abstract should be included or excluded based on the provided criteria and examples. Think step by step.

Return the result in JSON format.

Criteria:
{criteria}

Instances:
{instances}

Title:
{title}

Abstract:
{abstract}

Output Format:

```
{{
  "confidence": confidence_score (0 to 1),
  "decision": "include or exclude"
}}
```

Output:

**Table 4 Dynamic Few-Shot Evaluation Prompt.**

## Supplementary Information B

## Sensitivity Analysis of Confidence Score Thresholds

| | **0.7** | **Ratio** | **0.8** | **Ratio** | **0.9** | **Ratio** |
|---|---|---|---|---|---|---|
| *CD012233* | 0.4419 | 12.50% | 0.4598 | 19.23% | **<u>0.4615</u>** | 32.45% |
| *CD012768* | 0.6179 | 7.27% | **<u>0.6230</u>** | 10.00% | 0.6034 | 19.09% |
| *CD011977* | 0.6517 | 0.55% | **<u>0.6667</u>** | 2.19% | 0.6591 | 7.10% |
| *CD012069* | 0.3742 | 1.12% | 0.3760 | 3.56% | **<u>0.3873</u>** | 12.97% |
| *CD012551* | 0.4946 | 6.20% | 0.4974 | 8.83% | **<u>0.5000</u>** | 13.53% |
| *CD004414* | 0.7097 | 0.68% | 0.7097 | 1.36% | **<u>0.7333</u>** | 4.76% |
| *CD012661* | 0.6479 | 1.45% | 0.6537 | 3.05% | **<u>0.6555</u>** | 8.11% |
| *CD011431* | 0.5582 | 8.04% | 0.5714 | 11.15% | **<u>0.5791</u>** | 20.84% |
| *CD011420* | 0.6970 | 6.78% | **<u>0.7077</u>** | 8.47% | **<u>0.7077</u>** | 10.17% |
| *CD010772* | 0.3377 | 10.51% | 0.3333 | 13.56% | **<u>0.3467</u>** | 21.36% |

**Table 5 F1 Score Comparison under Different Confidence Score Thresholds.**

## Supplementary Information C

## Performance Metrics Comparison of Different Prompting Strategies

| | Zero-Shot | | | Chain-of-Thought | | | Few-Shot | | | Dynamic Few-Shot | | |
|---|---|---|---|---|---|---|---|---|---|---|---|---|
| | **Precision** | **Recall** | **Accuracy** | **Precision** | **Recall** | **Accuracy** | **Precision** | **Recall** | **Accuracy** | **Precision** | **Recall** | **Accuracy** |
| *CD012233* | 0.4762 | 0.2857 | 0.9135 | 0.5000 | 0.3143 | 0.9159 | 0.4333 | 0.3714 | 0.9062 | 0.3800 | 0.5429 | 0.8870 |
| *CD012768* | 0.4783 | 0.8250 | 0.6091 | 0.4783 | 0.8250 | 0.6091 | 0.4545 | 0.8750 | 0.5727 | 0.4578 | 0.9500 | 0.5727 |
| *CD011977* | 0.8571 | 0.5217 | 0.8579 | 0.8846 | 0.5000 | 0.8579 | 0.8571 | 0.5217 | 0.8579 | 0.6744 | 0.6304 | 0.8306 |
| *CD012069* | 0.2410 | 0.7619 | 0.7490 | 0.2458 | 0.7436 | 0.7584 | 0.2449 | 0.7875 | 0.7486 | 0.2412 | 0.8278 | 0.7357 |
| *CD012551* | 0.3182 | 0.5738 | 0.8102 | 0.3303 | 0.5902 | 0.8158 | 0.3495 | 0.5902 | 0.8271 | 0.3659 | 0.7377 | 0.8233 |
| *CD004414* | 0.6364 | 0.4667 | 0.9592 | 0.6364 | 0.4667 | 0.9592 | 0.6471 | 0.7333 | 0.9660 | 0.6875 | 0.7333 | 0.9694 |
| *CD012661* | 0.7586 | 0.3587 | 0.9581 | 0.7901 | 0.3478 | 0.9587 | 0.7634 | 0.3859 | 0.9593 | 0.6746 | 0.6196 | 0.9623 |
| *CD011431* | 0.8226 | 0.1868 | 0.7870 | 0.8065 | 0.1832 | 0.7852 | 0.7984 | 0.3626 | 0.8181 | 0.7039 | 0.4615 | 0.8172 |
| *CD011420* | 0.8421 | 0.4211 | 0.8941 | 0.8333 | 0.3947 | 0.8898 | 0.8182 | 0.4737 | 0.8983 | 0.8214 | 0.6053 | 0.9153 |
| *CD010772* | 0.5294 | 0.2143 | 0.8610 | 0.5000 | 0.2143 | 0.8576 | 0.4762 | 0.2381 | 0.8542 | 0.3871 | 0.2857 | 0.8339 |

**Table 6 Precision, Recall, and Accuracy for Zero-Shot, Chain-of-Thought, Few-Shot, and Dynamic Few-Shot.**